\documentclass{article}

\usepackage{caption}

\newcommand{\insmat}[1]{\mathop{\mbox{\rm #1}}}
\newcommand{\subjectto}{\insmat{s.t.}}

\newcommand{\minimize} {\insmat{min}}
\newcommand{\ba}{\begin{array}}
\newcommand{\ea}{\end{array}}
\newcommand{\ds}{\displaystyle}
\newcommand{\myproblem}[4]
    {\ba{lc} {\ds #4_{#1}} & {#2} \\*
        {\subjectto} & {\ba[t]{c} #3 \ea}
     \ea}
\newcommand{\probmin}[3]
    {\begin{equation} \myproblem{#1}{#2}{#3}{\minimize}
         \end{equation}}

\begin{document}

\title{Electre Tri-Machine Learning Approach to the Record Linkage Problem}

\author{Renato De Leone%
        \thanks{School of Science and Technology,
        University of Camerino, Italy
        \textit{renato.deleone@unicam.it}}
        \and Valentina Minnetti%
        \thanks{Faculty of Information Engineering, Informatics and Statistics, Sapienza University of Rome, Italy  \textit{valentina.minnetti@uniroma1.it}  }
        \thanks{Italian National Institute of Statistics, Rome, Italy
        \textit{minnetti@istat.it}  }
}


\maketitle

\begin{abstract}
In this short paper, the Electre Tri-Machine Learning Method, generally used to solve ordinal classification problems, is proposed for solving the Record Linkage problem. Preliminary experimental results show that, using the Electre Tri method, high accuracy can be achieved and more than 99\% of the matches and nonmatches were correctly identified by the procedure.
\end{abstract}


\section{Introduction}
\label{sec:1}

Machine Learning is a scientific discipline that is concerned with the design and development of algorithms that allow computers to ``learn data''. More precisely, ``learn'' is here intended as the possibility to automatically recognize complex patterns and make ``intelligent'' decisions, based on information data. Hence, machine learning is closely related to fields such as statistics, probability theory, data mining, pattern recognition, artificial intelligence, adaptive control and theoretical computer science.

Machine learning algorithms can be classified in the following types:
\begin{itemize}
\item supervised learning algorithms: a function/classifier is generated, that maps outputs on the training inputs, based on labeled examples input-output;
\item unsupervised learning algorithms: patterns in the input are recognized, the  examples have no labels;
\item semi-supervised learning algorithms: supervised and unsupervised learning information is combined;
\item reinforcement learning: actions from observation of the world are generated. Every action has some impact in the environment and the environment provides feedbacks that are translated into a score that guide the learning process.
\end{itemize}
The principal supervised learning techniques  currently applied or under consideration at statistical agencies worldwide to solve the record linkage matching problem are: classification tree \cite{Cohen98, elfe2002}, support vector machine \cite{Bilenko2003, Christen2008a, Christen2008b} and neural network \cite{Wilson2011}.
In this short paper, another machine learning technique is proposed to solve the record linkage problem: the multi-criteria classification method Electre Tri.
It is the first time that multi-criteria machine learning technique is used to solve the record linkage problem.

This application answers to one of ``many challenges in applying supervised machine learning to record linkage matching'' \cite{Kenn15}, showing that the use of multi-criteria classification method Electre Tri to solve the record linkage problem provides good results in term of classification model performances.
The importance of this application is in light of the increasing development of the use of administrative sources data. In this context, an important problem is that of finding matching pairs of records from heterogeneous databases, while maintaining privacy of the databases parties. To this purpose secure computation of distance metrics is important for secure record linkage \cite{RCF2004}.

The  paper is organized as follows. Section \ref{sec:2} describes an introduction to the Record Linkage problem; then the next Section \ref{sec:3} describes the method Electre Tri, used to solved the Record Linkage and in the last Section \ref{sec:4} a preliminary experiment is conducted on simulated data. The paper closes with some final remarks and conclusions.


\section{Linked Data: the Record Linkage}
\label{sec:2}

Generally speaking, in integration of two data sets the objective is the detection of those records, in the different data sets, that belong to the same statistical unit. This action allows the reconstruction of a unique record of data that contains all the unit information collected from different data sources on that unit.

Therefore, record linkage is the methodology of bringing together corresponding records from two or more files or finding duplicates within files \cite{Winkler99}. In the first situation, the definition of record linkage in \cite{fellegi69} is more precise ``Record linkage is a solution to the problem of recognizing those records in two files which represent identical persons, objects, or events (said to be matched)''

The term record linkage originated in the public health area when files of individual patients were brought together using name, date-of-birth and other information \cite{Winkler99}.

One of the main motivations for the utilize of the record linkage method is the construction of the big data bases for answer to the new informative needs \cite{felle97}.
\\ In order to better understand the problem,  small practical example is now presented. Suppose the user wants to link two datasets of persons A and B, whose the variables Name, Address and Age are known.
\\ Suppose that Table \ref{tab:1} contains the following values:
\begin{center}
\captionof{table}{Data in the first dataset}
\label{tab:1}       
\begin{tabular}{|cccc|} \hline
Unit & Name & Address  &  Age  \\ \hline
a1 & John A Smith  & 16 Main Street  & 16  \\
a2 & Javier Martinez  & 49 E Applecross Road  &   33  \\
a3 & Gillian Jones   & 645 Reading Aev   & 22   \\
\hline
\end{tabular}
\end{center}
\vskip 1em
\noindent Furthermore, suppose that Table \ref{tab:2}  contains the following values:
\begin{center}
\captionof{table}{Data in the second dataset} \label{tab:2}       
\begin{tabular}{|cccc|} \hline
Unit & Name & Address  &  Age  \\ \hline
b1 & J H Smith  & 16 Main St  & 17  \\
b2 & Haveir Marteenez  & 49 Aplecross Raod &   36  \\
b3 & Jilliam Brown   &  123 Norcross Blvd   &   43   \\
\hline
\end{tabular}
\end{center}
The matching table $ A \times B $ contains two units referring probably to the same persons, that the method should individuate as matches: 'John A Smith' with 'J H Smith' and 'Javier Martinez' with 'Haveir Marteenez'.

Modern record linkage begins with the pioneering work of Newcombe et al. \cite{newcombe59}, who introduced odds ratio of frequencies and the decision rules for delineating matches and nonmatches. In recent years, advances have yielded computer system that incorporate sophisticated ideas from computer sciences, statistics and operational research \cite{Winkler99}.

Then, Fellegi and Sunter \cite{fellegi69} introduced a mathematical foundation for record linkage. Their theory demonstrated the optimality of the decision rules used by Newcombe and introduced a variety of ways of estimating crucial matching probabilities (parameters) directly from the files being matches.

Formally, given two files A and B to be matched, each pair $ (a,b) \in \Gamma = A \times B $ has to be classified into \textit{true match} or \textit{true nonmatch}.
\\ The odds ratios of probabilities is:
$$ R= \frac{Pr( \gamma \in \Gamma \mid M)}{ Pr( \gamma \in \Gamma \mid U) } $$
where  $ \gamma $ is an arbitrary agreement pattern in the comparison space   $ \Gamma $, $M$ is the set of of \textit{true matches} and $U$ is the set of \textit{true nonmatches}. Between these two sets, the intermediate set of the possible matches exists.
\\ The decision rule reported below helps to classify the pairs:
\begin{itemize}
\item if $ R > Upper $, then the pair $ (a,b) $ is a \textit{designated match},
\item if $ Lower \leq R \le Upper $, then the pair $ (a,b) $ is a \textit{designated potential match},
\item if $ R < Lower $, then the pair $ (a,b) $ is a \textit{designated nonmatch}.
\end{itemize}
The estimation of the thresholds Upper and Lower is not easy in an objective way; the choice is competence of the analyst.
In the decision rule, three different sets were created: the \textit{designated matches}, \textit{designated potential matches}, \textit{designated nonmatches}. They constitute the partition of the set of all the records in the space  $ \Gamma $ in three subsets   $ C_{3} $  (\textit{matches}),   $ C_{2} $ (\textit{potential matches}) and    $ C_{1} $  (\textit{nonmatches}), whose intersections are empty sets.

The idea is to solve the record linkage problem as a multi-criteria based classification problem, whose a priori defined classes are the subsets of the partition.

Without going into too much details, in the next section a brief introduction to the method Electre Tri is presented.


\section{The multi-criteria method Electre Tri}
\label{sec:3}

In Multi Criteria Decision Aid, a finite set of objects (alternatives, actions, projects) is evaluated by a finite set of criteria, which measure their performances.
A criterion is the real-valued function    $ g_{j} : A \rightarrow   \Re  $, such that   $ g_{j}(a_{k}) $ indicates the performance of the alternative   $ a_{k} $ on the criterion   $ g_{j} $. The comparison of any pair of alternatives   $ a_{i} $ and   $ a_{k} $ may be grounded to the comparison of the two values   $ g_{j}(a_{i}) $  and   $ g_{j}(a_{k}) $   \cite{Electre.manual}.
\\ In general, a criterion can be either of gain or cost type; gain means that the DM prefers the highest value, while cost means that the DM prefers the lowest value on the criterion.

Many types of criterion were studied in literature, such as true-criterion, pseudo-criterion, pre-criterion, semi-criterion and other types \cite{Electre.manual}.
\\ In the case of true-criterion, if the difference between two performances is positive, then the true-criterion structure implies that the alternatives are in the strict preference relation; while if the difference is equal to 0, then they are in indifference relation.

The Electre Tri is a pseudo-criterion-based method. This type of criterion takes into account that data can be affected by errors from uncertainty, imprecision and small differences or big can not imply the same binary relations. Small and big differences of performances have to imply different binary relations. To define ``small'' and ``big'', two values are considered, which are the preference and indifference thresholds.

In literature, grouping problems can be divided in clustering, classification and sorting problems, depending on the a \textit{priori/posteriori} knowledge of classes. The sorting problem is a classification problem, dealt with multi-criteria approach, requiring to Decision Maker (DM) any \textit{preference information}. So, the aim of an ordinal sorting problem consists in assigning each alternative in one of the ordered predefined categories.
\\ Formally, given $p$ predefined ordered categories  $ C_{1}, C_{2}, \dots, C_{p}$ and a finite set of $n$ alternatives   $A = \{a_{1}, a_{2}, \dots, a_{n} \}$, evaluated on a finite set of $ m $ criteria   $ G = \{  g_{1}, g_{2}, \dots, g_{m} \} $, in the case all criteria are gain-type, the relations among the categories are   $ C_{1} \prec  C_{2} \prec  \dots  \prec  C_{p} $, such that   $ b_{h}$ is the profile, upper limit of category   $C_{h}$ and lower limit of category   $C_{h+1}$. In this way,   $C_{1}$ and   $C_{p}$ are the worst and the best categories respectively.

The Electre Tri method is based on outranking relations, indicated with S, which characterize how the alternatives are compared with the profiles.  Because the assignment of an alternative to a specific category follows from the comparison, on all criteria, of its performances with the profiles ones.
\\ The relation  $aSb_{h}$ validates or invalidates the assertion ``$a$ \textit{outranks} $b_{h}$'' whose meaning is  ``$a$ is at least as good as $b_{h}$'', on the set G.

In the context of the Electre Tri method, the validation of outranking relation is made by the computation of four indices \cite{Mousseau.Slowinski:1998, Electre.manual}:
\begin{enumerate}
\item the partial concordance indices on each criterion;
\item the global concordance index on all the criteria;
\item the partial discordance indices on each criterion;
\item the credibility index on all the criteria.
\end{enumerate}
For the computation of the partial concordance indices, it is necessary to know the profiles, preference and indifference thresholds values. In the case one of these parameters are not known, the index can not be computed. For the computation of the global concordance index is necessary to know the weights, representing the importance coefficients of the criteria. For the computation of the partial discordance indices, it is necessary to know the profiles, preference and veto thresholds values. And the credibility index corresponds to the global concordance index weakened by veto effects. If veto thresholds do not enter in the model,  the credibility index is equal to the global concordance index. From the credibility index to the definition of an outranking relation, it is necessary to fix a cutting level lambda, which is the minimum credibility index value which permits to define the outranking relation. Finally, the assignment of an alternative to one category does not result from the outranking relation directly, but it is necessary to use one (or both) of the two proposed exploitation procedures. They are the pessimistic and the optimistic assignment procedures. These procedures analyze the way an alternative compares to the profiles so as to determine the category to which the alternative should be assigned.

One of the main difficulties is the elicitation of various parameters that in the Electre Tri are profiles, weights, thresholds (preference, indifference and veto) and cutting level lambda. Even if these parameters can be interpreted, it can be difficult to fix directly their values (\textit{direct elicitation}) and to have a clear global understanding of the implications of these values in terms of the output \cite{Mousseau.Slowinski:1998}.
\\ In order to estimate indirectly the value of the parameters, De Leone and Minnetti  \cite{rdlvm:electre} proposed new estimation methodology whose procedure is composed of two phases: the first dedicated to the profiles and thresholds estimations, the second to the weights and cutting level estimations.
The core of the procedure is the profiles' estimation, suggested with Linear Programming (LP) using training set.

Let $p$ be the number of categories,   $m$ the number of criteria, the LP problem is the following:
\probmin{}{\hspace*{-5em} \ds \sum_{j=1}^{m} \sum_{a_{k}\rightarrow C_{h}}  \theta_{j}(a_{k}) }
{
\begin{array}{ll}
\theta_{j}(a_{k}) \geq g_{j}(a_{k}) - g_{j}(b_{h})  & \forall j = 1, \dots, m, \forall a_{k} \rightarrow C_{h, h \neq p}      \\
\theta_{j}(a_{k}) \geq g_{j}(b_{h-1}) - g_{j}(a_{k})   & \forall j = 1, \dots , m ,\forall a_{k} \rightarrow C_{h, h \neq 1 }     \\
g_{j}(b_{h}) \ge g_{j}(b_{h-1}) + \epsilon  & \forall j = 1, \dots , m , \forall h=2, \dots ,p-1      \\
\theta_{j}(a_{k}) \geq 0  & \forall j=1, \dots , m ,\forall a_{k} \rightarrow C_{h}
\end{array}}

 where   $ \epsilon $ is a small positive value.
\\ The problem (1) minimizes the sum of the classification errors   $ \theta_{j}(a_{k}) $ on all criteria and on all the alternatives in the training set, when this alternative's performance lies out the belonged category. The first two constraints define the error    $ \theta_{j}(a_{k}) $.

\section{Application to Real Data: a first experiment}
\label{sec:4}

As said in the previous section, the multi-criteria approach requires DM any \textit{preference information}, including binary relations. Since it is possible to state binary relations between the subsets as
$ C_{3}  \succ  C_{2}   \succ   C_{1}   $, the record linkage problem can be structured as ordinal sorting problem, that is, classification problem whose classes are ordered in the strict preference binary relations.

Moreover, the importance of using multi-criteria decision methods, with respect to the other classification methods, is in  the possibility to assign weights to each criterion, not possible in all the classification problems, and to use the \textit{preference information}, provided by DM, for estimating the classification model's parameters.

The proposed application wants to find a classification model (i.e. classifier or learner), assigning each record  of the space   $ \Gamma $ to one of the three categories   $ C_{1} $,   $ C_{2} $ and   $ C_{3} $, following the two phases procedure formulated by De Leone and Minnetti \cite{rdlvm:electre}.

The input data, used in the application, were taken from Winkler from American Census (in SecondString file for approximate string matching techniques).
Two data sets A and B are considered, containing 449 and 392 records respectively, and the true links are 327.

The variables (textual fields from synthetic census data) are the following:
\begin{itemize}
\item DS (\textit{labels of the data sets with A and B});
\item IDENTIFIER;
\item SURNAME;
\item NAME;
\item LASTCODE (\textit{middle name initial});
\item NUMCODE (\textit{address street number});
\item STREET (\textit{address street name}).
\end{itemize}
In this short paper, results from preliminary experiments are reported, because the application is an ongoing research, due to its complexity.

Some variables contain missing values that cause difficulties in the analysis, making it more complicated. So in order to facilitate the analysis, the records with missing values are deleted.

There are a number of popular methods of estimating the learner's ability to generalize; the test set method was used here.
In this experiment, the use of distance measure and the search of training set had played the most important roles; they had contributed to obtain good results of the classification model, found by Electre Tri \cite{Min14}.

The performance of the classification model, applied to the test set (83868 alternatives) was 99.09\% when all the criteria have the same importance and the lambda parameter is   $ \lambda = 0.50 $.
If lambda increases, the performance increases, up to 99.81\% when   $ \lambda = 0.70 $ and 99.89\% when   $ \lambda = 0.85 $.
\\ In the case the importance coefficients of criteria were considered different, the performances of the models were substantially the same, varying the lambda parameter.
\\ In the case of performance 99.09\%, the classification errors were committed by the model on the false links; namely, the model saw almost all the true links. The opposite situation occurred in the case of performance 99.89\%, when the model saw almost all the false links and misclassified the true links. To the DM the choice of the most interesting model, depending his preferences.

\section{Final Conclusion and Remarks}
\label{sec:5}

In this short paper, the Electre Tri machine learning technique was proposed for solving Record Linkage matching. It is the first time that multi-criteria decision technique is used to solve the record linkage problem.

The proposed application started with an initial experiment demonstrating that the application of the Electre Tri to record linkage shall provide good results in terms of classifier performances.
This paper shows only the results of a preliminary experiment, which provided good results in terms of performances of the classification model. Also this experiment confirmed that record linkage is more sensitive to the quality of preprocessing and standardization that of matching, as said in \cite{Win14}.

As consequence, other measures of distance in the construction of the input data matrix, as well as, different schemes in the search of training set, will be used.

\bibliographystyle{plain}
\bibliography{biblio}

\begin{thebibliography}{10}

\bibitem{Bilenko2003}
M.~Bilenko and R.~Mooney.
\newblock Adaptive duplicate detection using learnable string similarity.
\newblock {\em ACM SIGKDD}, pages 39--48, 2003.

\bibitem{Christen2008a}
P.~Christen.
\newblock Automatic record linkage using seeded nearest neighbour and support
  vector machine classification.
\newblock {\em ACM SIGKDD}, pages 151--159, 2008.

\bibitem{Christen2008b}
P.~Christen.
\newblock Automatic training example selection for scalable unsupervised record
  linkage.
\newblock {\em PAKDD, Springer LNAI}, 5012:511--518, 2008.

\bibitem{Cohen98}
W.W. Cohen.
\newblock The whirl approach to data integration.
\newblock {\em IEEE Intelligent Systems}, 13,no.3:20--24, 1998.

\bibitem{RCF2004}
W.W. Cohen, P.~Ravikumar, and S.E. Fienberg.
\newblock A secure protocol for computing string distance metrics.
\newblock In {\em PSDM held at ICDM}, pages 40--46, 2004.

\bibitem{rdlvm:electre}
R.~De~Leone and V.~Minnetti.
\newblock The estimation of the parameters in multi-criteria classification
  problem: The case of the electre tri method.
\newblock In Donatella Vicari, Akinori Okada, Giancarlo Ragozini, and Claus
  Weihs, editors, {\em Analysis and Modeling of Complex Data in Behavioral and
  Social Sciences}, Studies in Classification, Data Analysis, and Knowledge
  Organization, pages 93--101. Springer International Publishing, 2014.
\newblock {ISBN:} 978-3-319-06691-2.

\bibitem{elfe2002}
M.~Elfeky and A.~Elmagarmid.
\newblock Tailor: A record linkage toolbox.
\newblock {\em IEEE ICDE}, pages 17--28, 2002.

\bibitem{felle97}
I.P. Fellegi.
\newblock Record linkage and public policy: a dynamic evolution.
\newblock In W.~Alvey and B.~Jamerson, editors, {\em Proceedings of
  International Workshop and Exposition}. Record Linkage Techniques, 1997,
  March, 1997.

\bibitem{fellegi69}
I.P. Fellegi and A.B. Sunter.
\newblock A theory for record linkage.
\newblock {\em Journal of the American Statistical Association}, 64:1183--1210,
  1969.

\bibitem{Kenn15}
C.~Kenneth and C.~Poirier.
\newblock Machine learning documentation initiative.
\newblock {\em Modernisation Committee on Production and Methods, Statistics
  Canada}, Febrary 4, 2015.

\bibitem{Min14}
V.~Minnetti.
\newblock {\em On the parameters of the Electre Tri method: a proposal of a new
  two phases procedure}.
\newblock PhD thesis, Faculty of Information Engineering, Informatics and
  Statistics, Sapienza University of Rome, March 2015.

\bibitem{Mousseau.Slowinski:1998}
V.~Mousseau and R.~Slowinski.
\newblock Inferring an electre tri model from assignment examples.
\newblock {\em Journal of Global Optimization}, 12:157--174, March 1998.

\bibitem{Electre.manual}
V.~Mousseau, R.~Slowinski, and P.~Zielniewicz.
\newblock {\em ELECTRE TRI 2.0, a methodological guide and user\'s manual}.
\newblock Document du LAMSADE no111, Université Paris-Dauphine, 1999.

\bibitem{newcombe59}
H.B. Newcombe, J.M. Kennedy, S.J. Axford, and A.P. James.
\newblock Automatic linkage of vital records.
\newblock {\em Science}, 130(3381):954--959, October 1959.

\bibitem{Wilson2011}
D.R. Wilson.
\newblock Beyond probabilistic record linkage: using neural network and complex
  features to improve genealogical record linkage.
\newblock In {\em Proceedings of International Joint Conference on Neural
  Networks}, San Jose, California, USA, 2011.

\bibitem{Winkler99}
W.E. Winkler.
\newblock The state of record linkage and current research problems.
\newblock Technical report, U.S. Bureau of the Census, Statistics of Income
  Division, Internal Revenue Service Publication, R99/04, 1999.

\bibitem{Win14}
W.E. Winkler.
\newblock Matching and record linkage.
\newblock {\em WIREs Comput Stat}, 6:313--325, 2014.

\end{thebibliography}
\end{document}